\documentclass[letterpaper, 10 pt, conference]{ieeeconf}  

\usepackage{amsmath}
\usepackage{amssymb}
\usepackage{cite}
\let\labelindent\relax
\usepackage{enumitem}
\usepackage{hyperref}
\hypersetup{hidelinks,colorlinks=true,allcolors=black,pdfstartview=Fit,breaklinks=true}
\usepackage{graphicx}

\IEEEoverridecommandlockouts                              

\title{\LARGE \bf
Robust Distributed Control within a Curve Virtual Tube for a Robotic Swarm under Self-Localization Drift and Precise Relative Navigation
}

\author{Yan Gao, Chenggang Bai, Quan Quan
\thanks{Yan Gao, Chenggang Bai and Quan Quan are with the School of Automation Science and Electrical Engineering, Beihang University, Beijing 100191, P. R. China (email: { buaa\_gaoyan@buaa.edu.cn}; {bcg@buaa.edu.cn}; {qq\_buaa@buaa.edu.cn}).}}

\begin{document}

\maketitle
\thispagestyle{empty}
\pagestyle{empty}

\begin{abstract}
To guide the movement of a robotic swarm in a corridor-like environment,  a curve virtual tube with no obstacle inside is designed in our previous work. This paper generalizes the controller design to the condition that all robots have self-localization drifts and precise relative navigation, where the flocking algorithm is introduced to reduce the negative impact of the self-localization drift. It is shown that the cohesion behavior and the velocity alignment behavior are able to reduce the influence of the position measurement drift and the velocity measurement error, respectively.  For the convenience in practical use, a modified vector field controller with five control terms is put forward. Finally, the effectiveness of the proposed method is validated by numerical simulations and real experiments.

\end{abstract}


\section{Introduction}
In recent years, there has been an increasing interest in robotic swarm systems operating in a corridor-like environment \cite{yang2021survey}. Corridor-like environments mentioned here refer to indoor narrow corridors, indoor openings (such as doorframes and windows), outdoor complex environments (such as forests and urban environments), narrow water channels, etc \cite{park2022online}. In this process, not only should each robot avoid collisions with obstacles, but all robots also need to avoid collisions with each other. 

To solve the problem of robots operating in a corridor-like environment, many well-designed methods have been proposed, which can be classified as follows: formation, multi-robot trajectory planning and control-based methods. Each robot in the \emph{formation} usually remains a prespecified pose and makes transformation operations \cite{zhao2019bearing,xu2020affine}. The main weakness of the formation is its limited scalability and adaptability. 
The \emph{multi-robot trajectory planning} produces collision-free trajectories for all robots with a higher-order continuity \cite{park2022online,Zhou(2020)}. However, due to the sharp increase in the computational complexity and communication pressure, trajectories planning may become infeasible when multiple robots operate densely \cite{Zhou(2020)}. The \emph{control-based methods} are widely used for the robotic swarm because of their simplicity and accessibility \cite{panagou2015distributed,panagou2014motion,Wang(2017)}, which are most appropriate for a large robotic swarm.


\begin{figure}[tbp]
	\centering
	\includegraphics[scale=0.5]{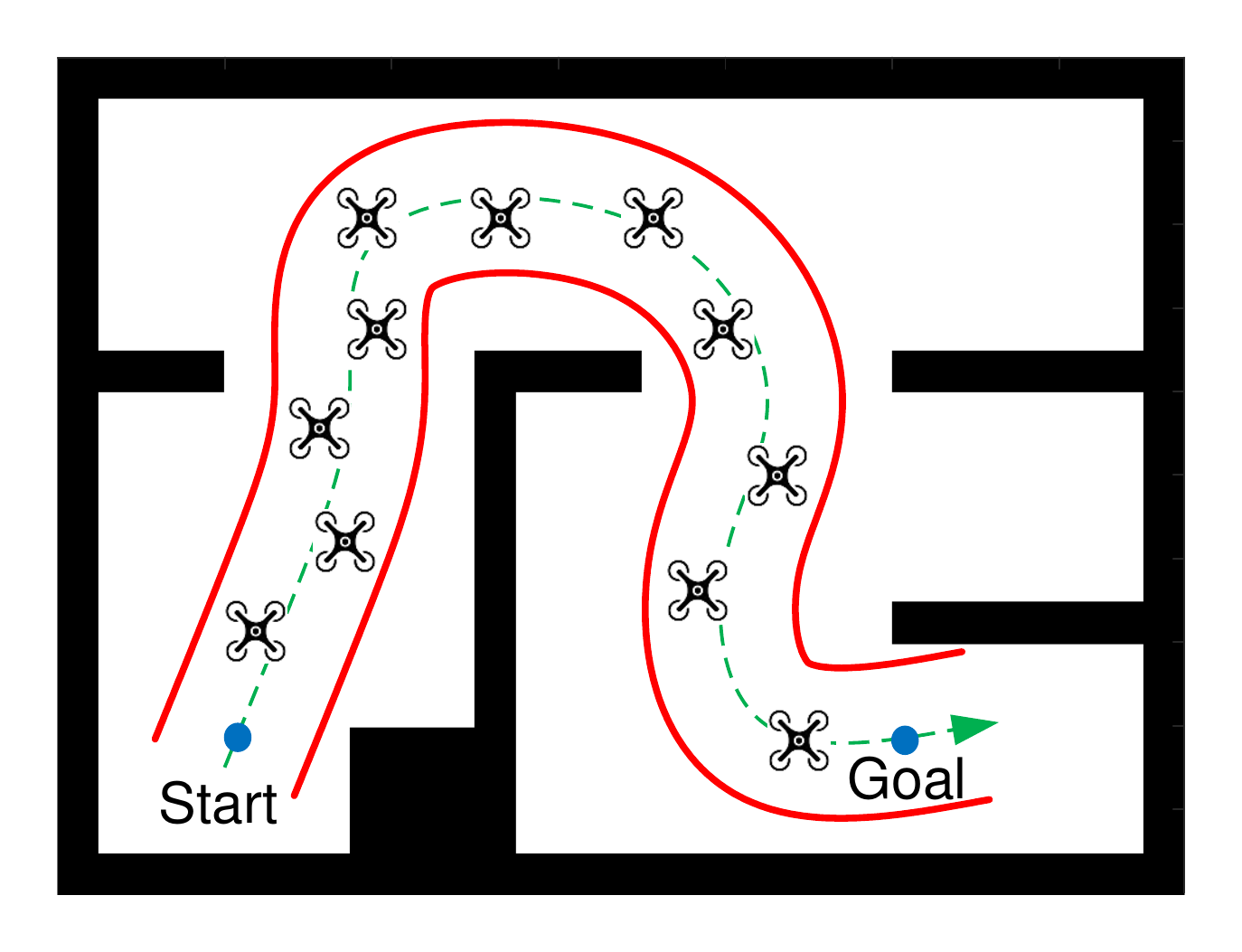}
	\caption{A curve virtual tube is designed to guide a quadcopter swarm in an indoor corridor-like environment.}
	\label{First}
\end{figure}

In our previous work \cite{quan2021distributed}, a \textit{curve virtual tube} is proposed for guiding the robotic swarm in a corridor-like environment. There is no obstacle inside the curve virtual tube, and robots only need to guarantee no collision with each other and the tube boundary. The term ``virtual tube" first appears in the AIRBUS’s Skyways project \cite{AIRBUS}. Besides,  as shown in Fig. \ref{First}, the concept of the curve virtual tube is similar to the \emph{lane} for autonomous road vehicles \cite{luo2018porca}, the \emph{safe flight corridor} for quadcopters \cite{park2022online}, and the multi-drone skyway framework CORRIDRONE \cite{Tony(2020)}. To guide and control a robotic swarm within a curve virtual tube, a distributed vector field controller is proposed in  \cite{quan2021distributed}, which is a type of control-based method.
However, there is no uncertainty considered in \cite{quan2021distributed}, namely all robots can obtain the information precisely and execute the command exactly. It is obvious that this assumption is too ideal for real practice. With regard to control-based methods, there exist some solutions to deal with uncertainties in the literature. In \cite{rezende2018robust,Rezende(2020)}, the authors analyze the influence of norm-bounded disturbances of a quadcopter or a fixed-wing UAV from the target curve. In \cite{olavo2018robust}, the authors present a robust guidance strategy to deal with additive perturbations representing norm-bounded uncertainties. In \cite{han2019robust}, the authors consider the robust multitask formation control problem for multiple agents with uncertain parameters. 

However, most of the studies on control-based methods only consider the influence of norm-bounded matched disturbances. The negative impact of the \emph{self-localization drift} is rarely considered, which includes \emph{position measurement drift} and \emph{velocity measurement error}. In many cases, robots operate in GPS-denied environments and have odometry systems, such as high-precision inertial odometry \cite{johnson2021development}, wheel odometry \cite{mohamed2019survey}, and visual-inertial odometry \cite{qin2018vins}, to get their ego-position and velocity information. It is well known that the position information obtained from state-of-the-art odometry systems is accurate in the short term, but drifts over time due to accumulated errors \cite{nguyen2021flexible}. The velocity information obtained from odometry systems is not precise, either. As a result, robots may have collisions with each other and have chaos in the swarm \cite{han2019robust}. Besides, position measurement drifts may also cause robots to reach wrong destinations \cite{nguyen2021flexible,zhang2022agile}.

In this paper, the \textit{robust curve virtual tube passing through problem} is summarized and solved. To achieve collision avoidance among robots, a traditional method is to share self-observation positions with the known initial positions among robots, which suffers heavily from the position measurement drift and communication uncertainties \cite{zhang2022agile}. The controller proposed in this paper can work autonomously without wireless communication and other robots' IDs, whose premise is that all robots have omnidirectional relative localization equipment to achieve precise relative navigation, namely robots can get their neighboring robots’ relative position and relative velocity precisely. The specific methods can be found in \cite{yan2019active,jayasuriya2020active,pavliv2021tracking}. Any robot has no need to identify other robots carefully, and the system structure is greatly simplified.
To modify the original controller in \cite{quan2021distributed}, three kinds of flocking behaviors of the Boids model \cite{reynolds1987flocks}, namely cohesion behavior, collision avoidance behavior, and velocity alignment behavior, are introduced. The controller in \cite{quan2021distributed} already includes a collision avoidance term for guaranteeing safety among robots. This paper shows that the cohesion behavior and the velocity alignment behavior can reduce the negative impact of the position measurement drift and the velocity measurement error, respectively. And a modified vector field controller is generated by adding two corresponding terms to the original
controller  in \cite{quan2021distributed}. However, it should be noted that this modified controller cannot always guarantee all robots to keep moving inside the curve virtual tube under the self-localization drift and precise relative navigation, which is obviously unrealistic when the position measurement drift is very large. 


The major contributions of this paper are summarized as follows.
\begin{itemize}[leftmargin=*]
	\item The flocking algorithm is introduced to reduce the negative impact of the self-localization drift when the robotic swarm is passing through a curve virtual tube.
	\item Formal proofs are proposed to show that the cohesion behavior and the velocity alignment behavior have the ability to reduce the negative impact of position measurement drift and velocity measurement error, respectively.
\end{itemize}

\section{Preliminaries and Problem Formulation}
\subsection{Robot Model}
\subsubsection{Robot Kinematics Model}
The robotic swarm consists of $M$ mobile robots in $\mathbb{R}^2$. Each robot has a double-integrator holonomic kinematics
\begin{align}
	\dot{\mathbf{{p}}}_{i} &=\mathbf{v}_{i} \label{FisrtOrder}\\
	\dot{\mathbf{{v}}}_{i} &=\mathbf{a}_{\text{c},i} \label{SecondOrder}, 
\end{align}
in which $\mathbf{a}_{\text{c},i}\in {{\mathbb{R}}^{2}}$ indicates the acceleration command of the $i$th robot, and  $\mathbf{p}_{i},\mathbf{v}_{i}\in {{\mathbb{R}}^{2}}$ stand for the position and velocity of the $i$th robot, respectively. The controller proposed in our previous work \cite{quan2021distributed} is a type of vector field controller. Hence, a hierarchical control architecture is necessary, namely first design the vector field $\mathbf{v}_{\text{c},i} \in {{\mathbb{R}}^{2}}$ , then design an acceleration command $\mathbf{a}_{\text{c},i}$ to make the robot track $\mathbf{v}_{\text{c},i}$. In the following, $\mathbf{v}_{\text{c},i}$ is also called as the \emph{velocity command}. 


\subsubsection{Four Areas around a Robot}



At the time $t>0$, the \emph{safety area} $\mathcal{S}_{i}$ of the $i$th robot is defined as
$
	\mathcal{S}_{i}\left(t\right)=\left \{ \mathbf{x}\in {{\mathbb{R}}^{2}}: \left \Vert \mathbf{x}-\mathbf{p}_{i}\left(t\right)\right \Vert \leq r_{\text{s}} \right \} ,
$
where $r_{\text{s}}>0$ is the \emph{safety radius}. For all robots, no \emph{conflict} with each other implies that $\mathcal{S}_{i}\cap \mathcal{S}_{j}=\varnothing$,
where $i,j=1,\cdots ,M,i\neq j$. The \emph{avoidance area} is defined for starting the collision avoidance control, which is expressed as  
$
	\mathcal{A}_{i}\left(t\right)=\left \{ \mathbf{x}\in {{\mathbb{R}}^{2}}:\left
	\Vert \mathbf{x}- \mathbf{p}_{i}\left(t\right)\right \Vert \leq r_{\text{a}}
	\right \},
$
and $r_{\text{a}}>0$ is the \emph{avoidance radius}. 
It is required that
$r_{\text{a}}>r_{\text{s}}$. The \emph{detection area} depends on the detection range of the relative localization equipment, which is shown as 
$
	\mathcal{D}_{i}\left(t\right)=\left \{ \mathbf{x}\in {{\mathbb{R}}^{2}}:\left
	\Vert \mathbf{x}- \mathbf{p}_{i}\left(t\right)\right \Vert \leq r_{\text{d}}
	\right \}  ,
$
and $r_{\text{d}}>0$ is the \emph{detection radius}. The set $\mathcal{N}_{\text{m},i}$ is defined as the collection of all mark numbers of other robots within $\mathcal{D}_{i}$, namely 
$
	\mathcal{N}_{\text{m},i}=\left \{ j: \left
	\Vert \mathbf{p}_{j}\left(t\right)- \mathbf{p}_{i}\left(t\right)\right \Vert \leq r_{\text{d}}\right \},
$
where $i\neq j$. Finally, at the time $t>0$, the \emph{cohesion area} $\mathcal{C}_{i}$ of the $i$th robot is defined as
$
	\mathcal{C}_{i}\left(t\right)=\left \{ \mathbf{x}\in {{\mathbb{R}}^{2}}: r_{\text{c}} \leq \left \Vert \mathbf{x}-\mathbf{p}_{i}\left(t\right)\right \Vert \leq r_{\text{d}} \right \} ,
$
where $r_{\text{c}}>0$ is the \emph{cohesion lower bound}. When the $j$th robot is within $\mathcal{C}_{i}$, namely $r_{\text{c}} \leq \left \Vert \mathbf{p}_{j}-\mathbf{p}_{i}\right \Vert \leq r_{\text{d}}$, the $i$th and $j$th robots have tendencies to move towards each other until their distance becomes smaller than $r_{\text{c}}$.

\subsubsection{Robot Observation Model}
In this paper, it is assumed that each robot has an odometry system to get its ego-position and velocity in its local coordinate system. The position measurement drift of the odometry system can be modeled as a random walk of the robot's local coordinate system \cite{nguyen2021flexible}. And the velocity measurement error can be modeled as a zero mean white Gaussian noise \cite{delaune2021range}. Hence, the robot observation model is expressed as 
\begin{align*}
	\dot{\hat{\mathbf{p}}}_{i} &=\mathbf{v}_{i}+\mathbf{n}_{\mathbf{p}_i} \\
	{\hat{\mathbf{v}}}_{i} &=\mathbf{v}_{i}+\mathbf{n}_{\mathbf{v }_i}, 
\end{align*}
where $\hat{\mathbf{p}}_{i},\hat{\mathbf{v}}_{i} \in {{\mathbb{R}}^{2}}$ represent the self-observation position and velocity of the $i$th robot, $\mathbf{n}_{\mathbf{p}_i} \sim \mathcal{N}\left(\mathbf{0},\sigma_{\mathbf{p}} \right)$ and $\mathbf{n}_{\mathbf{v}_i} \sim \mathcal{N}\left(\mathbf{0},\sigma_{\mathbf{v}} \right)$ indicate two kinds of two-dimensional zero mean white Gaussian noises and there exists $\sigma_{\mathbf{p}},\sigma_{\mathbf{v}}>0$. Suppose that there is no position measurement drift in the beginning. Then, at the time $t$, the relationship between the self-observation position and the real position of the $i$th robot is shown as \cite{nguyen2021flexible}
\begin{align}
	\hat{\mathbf{p}}_{i}\left(t\right) &={\mathbf{p}}_{i}\left(t\right)+\mathbf{r}_{\mathbf{p}_i}, \label{pihatpi}
\end{align}
where $\mathbf{r}_{\mathbf{p}_i} \sim \mathcal{N}\left(\mathbf{0},\sigma_{\mathbf{p}}\frac{t}{\Delta t} \right)$ and $\Delta t$ is the observation sampling time.

Define a position error $\tilde{\mathbf{p}}_{\text{m},ij}$ and a velocity error $\tilde{\mathbf{v}}_{\text{m},ij}$ between the $i$th and $j$th robots, which are shown as
$\tilde{\mathbf{p}}_{\text{m},ij} = \mathbf{p}_{i}-{{\mathbf{p}}_{j}},
	\tilde{\mathbf{v}}_{\text{m},ij} = \mathbf{v}_{i}-{{\mathbf{v}}_{j}}.
$
It is obvious that $\dot{\tilde{\mathbf{p}}}_{\text{m},ij}=\tilde{\mathbf{v}}_{\text{m},ij}$.
For precise relative navigation, an assumption is proposed as follows.

\textbf{Assumption 1}. Any robot is able to obtain relative position information and relative velocity information precisely. Meanwhile, the self-observation positions and velocities are not accurate, namely $\hat{\mathbf{p}}_{i},\hat{\mathbf{v}}_{i},\tilde{\mathbf{p}}_{\text{m},ij},\tilde{\mathbf{v}}_{\text{m},ij}, i,j=1,2,\cdots,M,i\neq j$ available.

%
%

\subsection{Curve Virtual Tube Model}

In our previous work \cite{quan2021distributed}, a \emph{curve virtual tube} is proposed for guiding a robotic swarm in a corridor-like environment. Here its definition and some necessary concepts are reviewed.

\begin{figure}[h]
	\begin{centering}
		\includegraphics[scale=1]{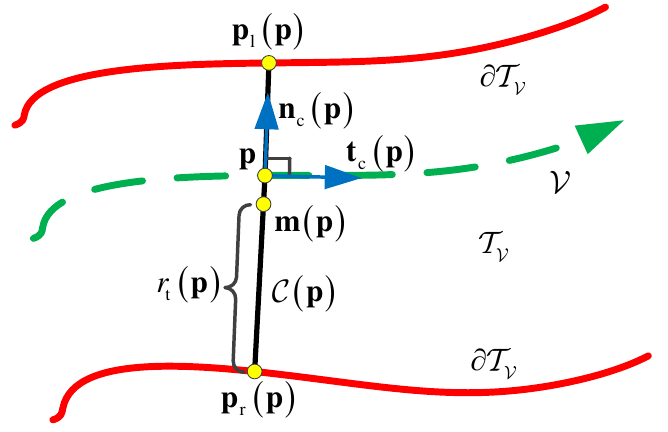}
		\par \end{centering}
	\caption{Relative concepts about the curve virtual tube.}
	\label{Curvetube}
\end{figure}

\begin{itemize}[leftmargin=*]
	\item \textbf{Generating Curve}. Inside the curve virtual tube, there exists a \emph{generating curve} $\mathcal{V} \subset \mathbb{R}^{2}$. As shown in Figure \ref{Curvetube}, if there exists $\mathbf{p}\in \mathcal{V}$, define $\mathbf{t}_{\text{c}}\left( 
	\mathbf{p}\right) \in \mathbb{R}^{2}$ to be the \emph{unit tangent vector} pointing in the forward direction, namely the moving direction. Similarly, $\mathbf{n}_{\text{c}}\left( \mathbf{p}\right) \in \mathbb{R}^{2}$ is
	the \emph{unit normal vector} directing anti-clockwise or left of the tangent direction. Then it has $\mathbf{t}_{\text{c}}^{\text{T}}\left( \mathbf{p}\right) \mathbf{n}_{\text{c}}\left( \mathbf{p}\right) \equiv 0.$
	
	\item \textbf{Cross Section}. For any $\mathbf{p}\in \mathcal{V},$ a \emph{cross section} passing $\mathbf{p}$ is defined as
	\begin{align*}
		\mathcal{C}\left(\mathbf{p}\right) =&\left \{  \mathbf{x}\in {{%
				\mathbb{R}}^{2}}: \mathbf{x}={{\mathbf{p}}}+\lambda
		\left( \mathbf{p}\right) \mathbf{n}_{\text{c}}\left( \mathbf{p}\right) ,\right. \\
		&\left. \lambda _{\text{l}}\left( \mathbf{p}\right) \leq
		\lambda \left( \mathbf{p}\right) \leq \lambda _{\text{r}}\left( \mathbf{p}%
		\right) ,\lambda _{\text{l}}\left( \mathbf{p}\right) ,\lambda _{\text{r}%
		}\left( \mathbf{p}\right) \in 	\mathbb{R}	\right \} .
	\end{align*}
	Here, $\mathcal{C}\left( {{\mathbf{p}}_{\text{f}}}\right)$ is called the \emph{finishing line} or \emph{finishing cross section}.
	For any point $\mathbf{p}^{\prime }\in \mathcal{C}\left( \mathbf{p}\right),$ $\mathcal{C}\left( \mathbf{p}^{\prime }\right) $ is defined as a cross section passing $\mathbf{p}^{\prime }$. It is obvious that $\mathcal{C}
	\left( \mathbf{p}^{\prime }\right) =\mathcal{C}\left( \mathbf{p}\right) .$ Besides, it has $\mathbf{t}_{\text{c}}\left( \mathbf{p}^{\prime }\right) =\mathbf{t}_{\text{c}}\left( 
	\mathbf{p}\right)$. 
	Two endpoints of $\mathcal{C}\left(\mathbf{p}\right)$ are defined as
$
		\mathbf{p}_{\text{l}}\left( \mathbf{p}\right)  \triangleq {{\mathbf{p}}}
		+\lambda _{\text{l}}\left( \mathbf{p}\right) \mathbf{n}_{\text{c}}\left( 
		\mathbf{p}\right),  
		\mathbf{p}_{\text{r}}\left( \mathbf{p}\right) \triangleq {{\mathbf{p}}}
		+\lambda _{\text{r}}\left( \mathbf{p}\right) \mathbf{n}_{\text{c}}\left( 
		\mathbf{p}\right) .
$
	The \emph{width} of  $\mathcal{C}\left( \mathbf{p}\right) $
	is expressed as 2$r_{\text{t}}\left( \mathbf{p}\right),$ which is defined as
$
		r_{\text{t}}\left( \mathbf{p}\right) \triangleq \frac{1}{2}\left \vert
		\lambda _{\text{r}}\left( \mathbf{p}\right) -\lambda _{\text{l}}\left( 
		\mathbf{p}\right) \right \vert .
$
	For any $\mathbf{p}^{\prime }\in \mathcal{C}\left(\mathbf{p}\right),$ the \emph{middle point} of $\mathcal{C}\left( \mathbf{p}\right)$  is defined as 
$
		\mathbf{m}\left( \mathbf{p}^{\prime }\right)=\mathbf{m}\left( \mathbf{p}\right)   \triangleq \frac{1}{2}\left( \mathbf{p}_{\text{l}}\left( \mathbf{p}\right)	+\mathbf{p}_{\text{r}}\left( \mathbf{p}\right) \right) .
$
	
	
	\item \textbf{Curve Virtual Tube}.  The curve virtual tube $\mathcal{T}_{\mathcal{V}}$ is generated by keeping cross sections always perpendicular to the tangent vectors of the given generating curve, which is denoted by
$
		\mathcal{T}_{\mathcal{V}}=\underset{\mathbf{p}\in \mathcal{V}}\cup \mathcal{C}\left( \mathbf{p}\right) .  
$
	Then the \emph{tube boundary} $\partial \mathcal{T}_{\mathcal{V}}$ is expressed as
	\begin{align*}
		\partial \mathcal{T}_{\mathcal{V}}=\left \{  \mathbf{x}\in {{%
				\mathbb{R}}^{2}}:
		\mathbf{x}=\mathbf{p}_{\text{l}}\left( \mathbf{p}\right) \cup \mathbf{p}_{\text{r}}\left( \mathbf{p}\right) ,\mathbf{p}\in \mathcal{V}\right \},
	\end{align*}
	which corresponds to two smooth boundary curves as shown in Figure \ref{Curvetube}.

	
\end{itemize}

%
%
%

\subsection{Distributed Controller for Passing through the Curve Virtual Tube under the Ideal Condition}
In \cite{quan2021distributed}, a distributed vector field controller is proposed for guiding the robotic swarm to pass through the curve virtual tube under the ideal condition.  The constant $v_{\text{m}}>0$ is set as the maximum permitted speed of all robots. Define a saturation function $\text{sat}\left( \mathbf{x},{v_{\text{m}}}\right)$ as 
\begin{equation*}
	\text{sat}\left( \mathbf{x},{v_{\text{m}}}\right) \triangleq
	\begin{cases}
		\mathbf{x}  & 	\left \Vert  \mathbf{x}\right \Vert \leq {v_{\text{m}}}\\ 
		{v_{\text{m}}}\frac{ \mathbf{x}}{\left \Vert  \mathbf{x}\right \Vert } & \left \Vert  \mathbf{x}\right \Vert  >{v_{\text{m}}}
	\end{cases}.
\end{equation*}
With barrier functions $V_{\text{m},ij}$, $V_{\text{t},i}$ in \cite{quan2021distributed} available, the controller for the $i$th robot is proposed as 
\begin{align}
	\mathbf{v}_{\text{c},i}&\!=\!\mathbf{v}\left( \mathcal{T}_{\mathcal{V}},\mathbf{p}_{i},
	\tilde{\mathbf{p}}{_{\text{m,}ij}}\right) \!=\!-\text{sat}\left(\mathbf{u}_{1,i}+\mathbf{u}_{2,i}+\mathbf{u}_{3,i},v_{\text{m}}\right),
	\label{modifiedcontroller}
\end{align}
where $\mathbf{u}_{1,i}$, $\mathbf{u}_{2,i}$, $\mathbf{u}_{3,i}$ are called as  \emph{line approaching term},  \emph{robot avoidance term} and  \emph{virtual tube keeping term}, respectively. These subcommands are shown as
$
	\mathbf{u}_{1,i}=-v_{\text{m}}\mathbf{t}_{\text{c}}\left( \mathbf{p}_{i}\right),
	\mathbf{u}_{2,i}=\sum_{j\in \mathcal{N}_{\text{m},i}}-b_{ij}\mathbf{\tilde{p}}_{\text{m},ij},
	\mathbf{u}_{3,i}=\left(\mathbf{I}_{2}-\mathbf{t}_{\text{c}}\left(  \mathbf{p}{_{i}}\right)\mathbf{t}_{\text{c}}^{\text{T}}\left(  \mathbf{p}{_{i}}\right) \right)\mathbf{c}_{i}.
$
The specific mathematical forms of these subcommands can be found in \cite{quan2021distributed}.

\begin{figure}[h]
	\begin{centering}
		\includegraphics[scale=1]{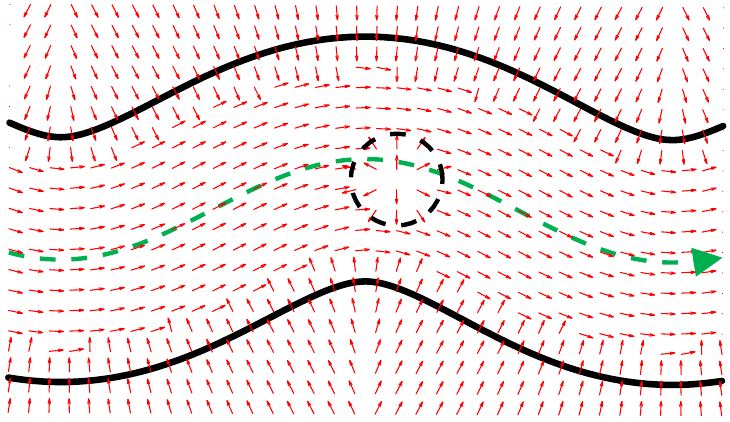}
		\par \end{centering}
	\caption{Vector field \eqref{modifiedcontroller} of a curve virtual tube.}
	\label{MVF}
\end{figure}

Consider a scenario that a robot is moving within a curve virtual tube, in the middle of which there exists another robot. Figure \ref{MVF} shows the vector field of this curve virtual tube with the swarm controller \eqref{modifiedcontroller}.

Then, a theorem is stated to show the system stability.

\textbf{Theorem 1} \cite{quan2021distributed}.  Under \textit{Assumptions 1-4} in \cite{quan2021distributed}, suppose that (i) all robots have a  single-integrator model $\dot{\mathbf{{p}}}_{i}=\mathbf{v}_{\text{c},i},i=1,\cdots ,M$; (ii) the vector field for the $i$th robot is designed as (\ref{modifiedcontroller}); (iii) if the $i$th robot has passed through the curve virtual tube $\mathcal{\mathcal{T}_{\mathcal{V}}}$, then $b_{ij}= 0$, $\mathbf{c}_{i}= \mathbf{0}$; (iv)  $\mathcal{\mathcal{T}_{\mathcal{V}}}$ is wide enough for at least one robot to pass through. Then, there exists $t_{1}>0 $ such that all robots can pass through  $\mathcal{T}_{\mathcal{V}}$ at $t\geq t_{1},$ meanwhile guaranteeing $\mathcal{S}_{i}\left(t\right)\cap \mathcal{S}_{j}\left(t\right)=\varnothing,$ $\mathcal{S}_{i}\left(t\right)\cap \partial \mathcal{T}_{\mathcal{V}}=\varnothing$, $t\in \lbrack 0,\infty ), i,j=1,2,\cdots,M,i \neq j$.

In this paper, the robot kinematics model \eqref{FisrtOrder}, \eqref{SecondOrder} is second-order. The objective now is to establish a control law for $\mathbf{a}_{\text{c},i}$ to make the robot track $\mathbf{v}_{\text{c},i}$ under the ideal condition. With $\mathbf{v}_{\text{c},i}$ and its derivative $\dot{\mathbf{v}}_{\text{c},i}$ available, the acceleration command for the $i$th robot is proposed as \cite{Rezende(2020)}
\begin{align}
	\mathbf{a}_{\text{c},i}&=\mathbf{a}\left(\mathbf{v}_{i},\mathbf{v}_{\text{c},i}\right)  =k_{\mathbf{v}}\left(\mathbf{v}_{\text{c},i}-\mathbf{v}_{i}\right)+\dot{\mathbf{v}}_{\text{c},i} \label{AccelerationController}
\end{align}
with $k_{\mathbf{v}}>0$. Then a lemma is put forward as follows.

\textbf{Lemma 1} \cite{Rezende(2020)}. When the vector field $\mathbf{v}_{\text{c},i}$ and its derivative $\dot{\mathbf{v}}_{\text{c},i}$ are available, there exist $\lim_{t\rightarrow\infty}\mathbf{v}_i\left(t\right)=\mathbf{v}_{\text{c},i}$ if the acceleration command is designed as \eqref{AccelerationController}.


\subsection{Problem Formulation}

With descriptions and preliminaries above, the \emph{robust curve virtual tube passing through problem} is stated as follows.

\textbf{Robust curve virtual tube passing through problem}. Under \textit{Assumption 1}, design the acceleration command $\mathbf{a}_{\text{c},i}$ to guide and control all robots to pass through the curve virtual tube $\mathcal{T}_{\mathcal{V}}$, meanwhile
avoiding colliding with each other ($\mathcal{S}_{i}\left(t\right)\cap \mathcal{S}_{j}\left(t\right)=\varnothing$) and keeping within the tube ($
\mathcal{S}_{i}\left(t\right)\cap \partial \mathcal{T}_{\mathcal{V}}=\varnothing$) as far as possible, where $i,j=1,\cdots ,M$, $i\neq j$, $t>0$.

\section{Robust Swarm Controller Design and Analysis}
\subsection{Robust Swarm Controller Design}
When the self-localization drift and precise relative navigation are considered, namely \textit{Assumption 1} is satisfied, the controllers \eqref{modifiedcontroller}, \eqref{AccelerationController} are rewritten as
\begin{align}
	\mathbf{v}_{\text{c},i}&=\mathbf{v}\left( \mathcal{T}_{\mathcal{V}},\hat{\mathbf{p}}_{i},
	\tilde{\mathbf{p}}{_{\text{m,}ij}}\right)  \!=\!-\text{sat}\left(\hat{\mathbf{u}}_{1,i}+\mathbf{u}_{2,i}+\hat{\mathbf{u}}_{3,i},v_{\text{m}}\right)\label{VelNoVA}\\
	\mathbf{a}_{\text{c},i}&=\mathbf{a}\left(\hat{\mathbf{v}}_{i},\mathbf{v}_{\text{c},i}\right) =k_{\mathbf{v}}\left(\mathbf{v}_{\text{c},i}-\hat{\mathbf{v}}_{i}\right)+\dot{\mathbf{v}}_{\text{c},i}\label{AccNoVA},
\end{align}
where
$
	\hat{\mathbf{u}}_{1,i}=-v_{\text{m}}\mathbf{t}_{\text{c}}\left( \hat{\mathbf{p}}_{i}\right),	\hat{\mathbf{u}}_{3,i}=\left(\mathbf{I}_{2}-\mathbf{t}_{\text{c}}\left(  \hat{\mathbf{p}}{_{i}}\right)\mathbf{t}_{\text{c}}^{\text{T}}\left(  \hat{\mathbf{p}}{_{i}}\right) \right)\mathbf{c}_{i}.
$
Obviously, position measurement drifts and velocity measurement errors have negative impacts on all robots. Robots may have collisions with each other and move outside the curve virtual tube. To solve this problem, the acceleration command is as same as the original controller \eqref{AccNoVA}, and the flocking algorithm is introduced into the modified vector field controller, which is shown as 
\begin{align}
	\mathbf{v}_{\text{c},i}&=\mathbf{v}_{\text{mdf}}\left( \mathcal{T}_{\mathcal{V}},\hat{\mathbf{p}}_{i},
	\tilde{\mathbf{p}}{_{\text{m,}ij}}\right) \label{vmdf}\\
	&=-\text{sat}\left(\hat{\mathbf{u}}_{1,i}+\mathbf{u}_{2,i}+\hat{\mathbf{u}}_{3,i}+\mathbf{u}_{4,i}+\mathbf{u}_{5,i},v_{\text{m}}\right), \notag
\end{align}
where the \emph{robot cohesion term} $\mathbf{u}_{4,i}$ and the \emph{velocity alignment term} $\mathbf{u}_{5,i}$ are shown as 
\begin{align*}
	\mathbf{u}_{4,i}&=\sum_{j\in \mathcal{N}_{\text{m},i}}\underset{d_{ij}}{\underbrace{\frac{\partial V_{\text{a},ij}}{\partial \left \Vert \tilde{\mathbf{{p}}}{_{\text{m,}ij}}\right \Vert }\frac{1}{\left \Vert \tilde{\mathbf{{p}}}{_{\text{m,}ij}}\right \Vert } }}\mathbf{\tilde{p}}_{\text{m},ij} =\sum_{j\in \mathcal{N}_{\text{m},i}}d_{ij}\mathbf{\tilde{p}}_{\text{m},ij}\\
	\mathbf{u}_{5,i}&=k_{5}\sum_{j\in \mathcal{N}_{\text{m},i}}\tilde{\mathbf{v}}_{\text{m},ij}
\end{align*}
with $k_{5}>0$. The attractive potential function $V_{\text{a},ij}$ in $\mathbf{u}_{4,i}$ will be introduced in the following.

\subsection{Analysis: Why Robot Cohesion Term Can Reduce Negative Impact of Position Measurement Drift}

\subsubsection{Attractive Potential Field Function for Robot Cohesion Term}
Before introducing the attractive potential function $V_{\text{a},ij}$, a smooth function $\delta \left(x,d_{1},d_{2}\right)$ is defined as
\begin{equation*}
	\delta \left(x,d_{1},d_{2}\right)=\left \{ 
	\begin{array}{c}
			0 \\ 
			Ax^{3}+Bx^{2}+Cx+D \\ 
			1
		\end{array}
	\right.
	\begin{array}{c}
			x\leq d_{1} \\ 
			d_{1}\leq x\leq d_{2} \\ 
			d_{2}\leq x%
		\end{array}
\end{equation*}
with $A=2\left/\left(d_{1}-d_{2}\right)^{3}\right.,$ $B=-3\left(d_{1}+d_{2}%
\right)\left/\left(d_{1}-d_{2}\right)^{3}\right.,$ $C=6d_{1}d_{2}\left/%
\left(d_{1}-d_{2}\right)^{3}\right.$, $D=d_{1}^{2}\left(d_{1}-3d_{2}%
\right)\left/\left(d_{1}-d_{2}\right)^{3}\right.$. 

\begin{figure}
	\centering
	\includegraphics[width=\columnwidth]{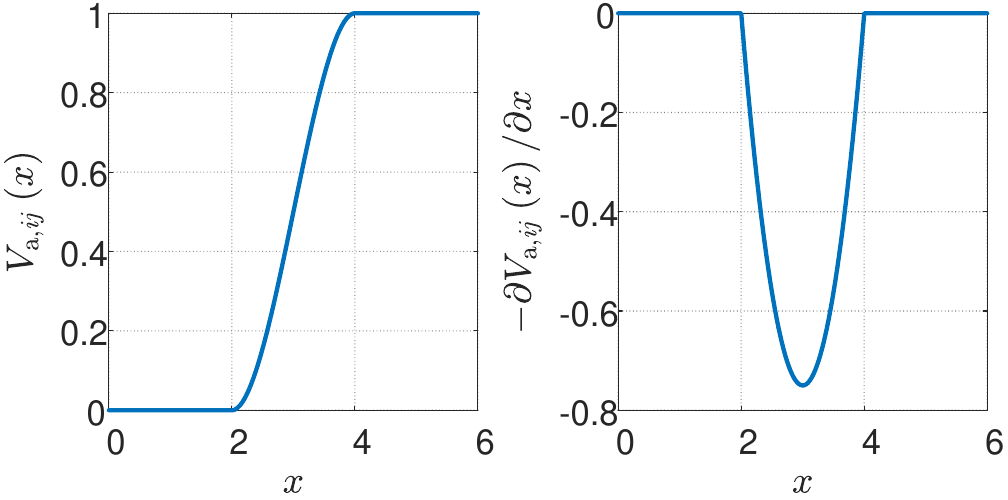}
	\caption{Attractive potential field function and its negative derivative.}
	\label{CohesionPlot}
\end{figure}

Then, according to the definition of the cohesion area $\mathcal{C}_{i}$, an attractive potential field function $V_{\text{a},ij}$ is defined as
\begin{equation}
	V_{\text{a},ij}=k_4\delta \left(\left \Vert \tilde{\mathbf{p}}{_{\text{m,}ij}}\right \Vert ,r_{\text{c}},r_{\text{d}}\right) \label{Vaij}
\end{equation}
with $k_4>0$. When there exist $k_4=1,r_{\text{c}}=2,r_{\text{d}}=4$, the function $V_{\text{a},ij}\left(x\right)=\delta \left(x,2,4\right)$ and its negative derivative $-\partial V_{\text{a},ij}\left(x\right) /\ \partial x$ are shown in Figure \ref{CohesionPlot}.
When $\left \Vert \tilde{\mathbf{p}}{_{\text{m,}ij}}\right \Vert<r_{\text{d}}$, the attractive forces of the $i$th and $j$th robots begin to appear. When $\left \Vert \tilde{\mathbf{p}}{_{\text{m,}ij}}\right \Vert<r_{\text{c}}$, the attractive forces disappear. Besides, when $\left \Vert \tilde{\mathbf{p}}{_{\text{m,}ij}}\right \Vert=\left(r_{\text{c}}+r_{\text{d}}\right)/2$, the attractive forces reach the maximum.

\textbf{Remark 1}. The introduction of the cohesion term has no influence on the system stability and safety. As $V_{\text{a},ij}$ is limited, this function is not a ``barrier'' function. Correspondingly, $\mathbf{u}_{4,i}$ in \eqref{vmdf} is always \emph{bounded}. Hence the cohesion behavior can be seen as a \emph{soft} constraint. On the contrary, the collision avoidance and the virtual tube keeping are two \emph{hard} constraints. When there exist safety risks of the $i$th robot, $\mathbf{u}_{2,i}$ or $\hat{\mathbf{u}}_{3,i}$ in \eqref{vmdf} will become \emph{unbounded}. Besides, when the distances between any pair of robots are all less than $r_\text{c}$ or larger than $r_\text{d}$, the controller \eqref{vmdf} will degenerate into \eqref{VelNoVA} with $\mathbf{u}_{5,i}$ unconsidered, namely the controller \eqref{vmdf} is still satisfied with \emph{Theorem 1} under the ideal condition.

\textbf{Remark 2}. Although $\mathbf{u}_{4,i}$ in \eqref{vmdf} can reduce the negative impact of the position measurement drift, the main cost is the moving efficiency reduction. The cohesion term makes robots move towards each other. And $\mathbf{u}_{4,i}$ of the robot in the front of the robotic swarm probably points to the opposite direction of $\mathbf{u}_{1,i}$. In other words, robots in the front have less force to move forward in $\mathcal{T}_{\mathcal{V}}$ due to the existence of $\mathbf{u}_{4,i}$. As a result, although robots in the rear have more force to move forward with the help of $\mathbf{u}_{4,i}$ from robots in the front, robots in the rear are blocked by other robots ahead. Therefore, the modified vector field controller \eqref{vmdf} reduces the moving efficiency in the curve virtual tube. 

\subsubsection{Effect Analysis of Robot Cohesion Term}
In the following, we are going to discuss why the robot cohesion term $\mathbf{u}_{4,i}$ in \eqref{vmdf} can reduce the negative impact of the position measurement error.

\begin{figure}[h]
	\centering
	\includegraphics[width=\columnwidth]{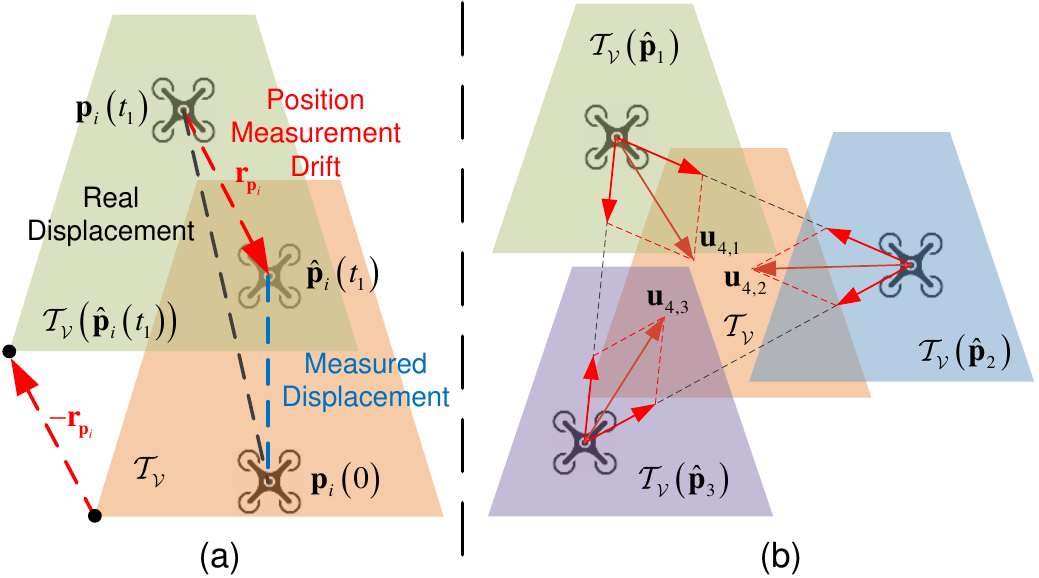}
	\caption{(a) The relationship between the position measurement drift and the drifting curve virtual tube. (b) An example of three robots having tendencies to move towards the curve virtual tube $\mathcal{T}_{\mathcal{V}}$ due to the existence of $\mathbf{u}_{4,i}$ in \eqref{vmdf}.}
	\label{PositionDrift3}
\end{figure}


As shown in Figure \ref{PositionDrift3}(a), at the time $t=0$, the $i$th robot locates at $\mathbf{p}_i\left(0\right)$ and begins to pass through $\mathcal{T}_{\mathcal{V}}$. Then, at the time $t=t_1$, its real position is $\mathbf{p}_i\left(t_1\right)$ and its measured position is $\hat{\mathbf{p}}_i\left(t_1\right)$. As $\hat{\mathbf{p}}_i\left(t_1\right)$ instead of ${\mathbf{p}}_i\left(t_1\right)$ is used in the controllers \eqref{VelNoVA} and \eqref{vmdf}, the $i$th robot is affected by the location relationship between $\hat{\mathbf{p}}_i\left(t_1\right)$ and  $\mathcal{T}_{\mathcal{V}}$, which can be considered equivalent to the $i$th robot keeping passing through its corresponding \emph{drifting curve virtual tube} $\mathcal{T}_{\mathcal{V}}\left(\hat{\mathbf{p}}_i\left(t_1\right)\right)$. This equivalence relationship actually transfers the position measurement drift from the robot to the curve virtual tube, namely there is no drift in the robot's position and the robot is affected by $\mathcal{T}_{\mathcal{V}}\left(\hat{\mathbf{p}}_i\left(t_1\right)\right)$.
Define the position of $\mathcal{T}_{\mathcal{V}}$ as $\mathbf{p}\left(\mathcal{T}_{\mathcal{V}}\right)$. Then, according to \eqref{pihatpi}, it has 
$
	\mathbf{p}\left(\mathcal{T}_{\mathcal{V}}\left(\hat{\mathbf{p}}_i\left(t_1\right)\right)\right)=\mathbf{p}\left(\mathcal{T}_{\mathcal{V}}\right)-\mathbf{r}_{\mathbf{p}_i}.
$
As the mean value of $\mathbf{r}_{\mathbf{p}_i}$ is a zero vector, $\mathcal{T}_{\mathcal{V}}\left(\hat{\mathbf{p}}_i\right)$ of all robots are evenly distributed around $\mathcal{T}_{\mathcal{V}}$. As stated in \textit{Remark 1}, the introduction of $\mathbf{u}_{4,i}$ has no influence on the system stability and safety. When there exist position measurement drifts, robots keep moving inside their corresponding $\mathcal{T}_{\mathcal{V}}\left(\hat{\mathbf{p}}_i\right)$, which results in that all robots are also evenly distributed around  $\mathcal{T}_{\mathcal{V}}$. Meanwhile, $\mathbf{u}_{4,i}$ in \eqref{vmdf} makes robots move toward each other. With relative properties of the consensus algorithm \cite{olfati2006flocking}, it is obvious that $\mathbf{u}_{4,i}$ directs to $\mathcal{T}_{\mathcal{V}}$ with a high probability, which increases with the number of robots inside $\mathcal{D}_{i}$. Figure \ref{PositionDrift3}(b) shows an example with three robots, whose $\mathcal{T}_{\mathcal{V}}\left(\hat{\mathbf{p}}_i\right)$ are evenly distributed around $\mathcal{T}_{\mathcal{V}}$. It can be easily observed that $\mathbf{u}_{4,i}, i=1,2,3$ all point to $\mathcal{T}_{\mathcal{V}}$. In a word, the robot cohesion term $\mathbf{u}_{4,i}$ in \eqref{vmdf} is able to make robots have more tendencies to keep within $\mathcal{T}_{\mathcal{V}}$. 


\subsection{Analysis: Why Velocity Alignment Term Can Reduce Negative Impact of Velocity Measurement Error}
\subsubsection{Modified Acceleration Command Containing a Velocity Alignment Term}
When the saturation function in \eqref{vmdf} does not work, the velocity alignment term $\mathbf{u}_{5,i}$ in \eqref{vmdf} can be directly transferred to the acceleration command. Hence, to facilitate the analysis, a modified acceleration command containing $\mathbf{u}_{5,i}$ is shown as
\begin{align}
	\mathbf{a}_{\text{c},i}=\mathbf{a}_{\text{mdf}}\left(\hat{\mathbf{v}}_{i},\mathbf{v}_{\text{c},i}\right)=k_{\mathbf{v}}\left(\mathbf{v}_{\text{c},i}-\hat{\mathbf{v}}_{i}-\mathbf{u}_{5,i}\right)+\dot{\mathbf{v}}_{\text{c},i}.  \label{ModifiedAccelerationController}
\end{align}

\subsubsection{Effect Analysis of  Velocity Alignment Term in the Modified Acceleration Command}
In real practice, the curvature change of $\mathcal{V}$ is quite small in most cases. Besides, compared with $r_\text{d}$, the length of $\mathcal{V}$ is much larger. Hence, $\mathbf{u}_{1,i}$ in \eqref{vmdf} of all robots in a detection area are almost the same. As we only discuss the relationship between velocity measurement errors and velocities of robots, it is assumed that all robots have the same zero velocity command, namely $\mathbf{v}_{\text{c},i}=\mathbf{0} $, which also results in $\dot{\mathbf{v}}_{\text{c},i}=\mathbf{0}, i=1,2,\cdots,M$.
Then an important lemma is stated as follows.


\textbf{Lemma 2}. Suppose that there are $N$ robots having relative localization relationship. The velocity $\mathbf{v}_{i}$ of the $i$th robot in \eqref{AccNoVA} and the one in \eqref{ModifiedAccelerationController} both contain a two-dimensional zero mean white Gaussian noise component, which is expressed as $\mathbf{n}_{\mathbf{v}_i}^{\prime} \sim \mathcal{N}\left(\mathbf{0},\sigma_{\mathbf{v}}^{\prime} \right)$ and $\mathbf{n}_{\mathbf{v}_i}^{\prime\prime} \sim \mathcal{N}\left(\mathbf{0},\sigma_{\mathbf{v}}^{\prime\prime} \right)$, respectively. Then there exist $\sigma_{\mathbf{v}}^{\prime}=\frac{k_\mathbf{v}}{2}\sigma_{\mathbf{v}}$ and $
\sigma_\mathbf{v}^{\prime\prime}=\frac{k_\mathbf{v}}{2}\frac{k_{5}+1}{k_{5}N+1}\sigma_{\mathbf{v}}$.


\textit{Proof}. See \emph{Appendix A}. $\square$ 

\textbf{Remark 3}. Compared with \eqref{AccNoVA}, the modified acceleration command \eqref{ModifiedAccelerationController} containing a velocity alignment term reduces the variance of the noise in the velocity. The variance ratio relates to $N$ and $k_{5}$, namely
$\frac{\sigma_\mathbf{v}^{\prime\prime}}{\sigma_\mathbf{v}^{\prime}}=\frac{k_{5}+1}{k_{5}N+1}.$ For any robot in the robotic swarm, it has $N=1$ when there is no other robot in its detection area. Otherwise, there exists $N>1$, which causes $\sigma_{v}^{\prime\prime} / \sigma_{v}^{\prime}<1$. 

\subsection{Analysis: Control Effectiveness of Robust Swarm Controller}

With \emph{Lemmas 1-2} and the above analyses available, the main result of this paper is stated as follows.

\textbf{Theorem 2}. Suppose that (i) all robots have a  double-integrator model \eqref{FisrtOrder}, \eqref{SecondOrder}; (ii) the controllers for the $i$th robot is designed as \eqref{AccNoVA}, \eqref{vmdf}; (iii) if the $i$th robot has passed through $\mathcal{\mathcal{T}_{\mathcal{V}}}$, then $b_{ij}= 0$, $\mathbf{c}_{i}= \mathbf{0}$; (iv)  $\mathcal{\mathcal{T}_{\mathcal{V}}}$ is wide enough for at least one robot to pass through.
Then, under the ideal condition, there exists $t_{1}>0 $ such that all robots can pass through  $\mathcal{T}_{\mathcal{V}}$ at $t\geq t_{1},$ meanwhile guaranteeing $\mathcal{S}_{i}\left(t\right)\cap \mathcal{S}_{j}\left(t\right)=\varnothing,$ $\mathcal{S}_{i}\left(t\right)\cap \partial \mathcal{T}_{\mathcal{V}}=\varnothing$, $t\in \lbrack 0,\infty ), i,j=1,2,\cdots,M,i \neq j$. And when \textit{Assumption 1} is satisfied, the controllers \eqref{AccNoVA}, \eqref{vmdf} can make the robotic swarm pass through $\mathcal{T}_{\mathcal{V}}$ more safely and more precisely compared with controllers \eqref{VelNoVA}, \eqref{AccNoVA}.

\textit{Proof}. See \emph{Appendix B}. $\square$

\section{Simulation and Experiment}
Simulations and experiments are given to show the effectiveness of the proposed method. A video is available on
\href{https://youtu.be/UjfkljQDElA}{https://youtu.be/UjfkljQDElA} and \href{http://rfly.buaa.edu.cn}{http://rfly.buaa.edu.cn}. 
\subsection{Comparative Numerical Simulations}
In this subsection, the validity and feasibility of the proposed method is numerically verified in comparative simulations under the self-localization drift and precise relative navigation. Consider a scenario that $M=6$ robots pass through a predefined curve virtual tube $\mathcal{T}_{\mathcal{V}}$. The width of $\mathcal{T}_{\mathcal{V}}$ is always $r_\text{t}=1\text{m}$ along the generating curve. All robots are arranged symmetrically in a rectangular space in the beginning and satisfy the second-order model in \eqref{FisrtOrder}, \eqref{SecondOrder}. The original controllers \eqref{VelNoVA}, \eqref{AccNoVA} and modified controllers  \eqref{AccNoVA}, \eqref{vmdf} are applied to guide these robots, respectively. The control parameters are $k_2=k_3=k_5=1$, $k_4=2$, $r_\text{s} = 0.2\text{m}$, $r_\text{a} = 0.3\text{m}$, $r_\text{c} = 1\text{m}$, $r_\text{d} = 2\text{m}$, $v_\text{m} = 1\text{m/s}$. The variances of noises $\mathbf{n}_{\mathbf{p}_i}$ and $\mathbf{n}_{\mathbf{v}_i}$ are set as $\sigma_{\mathbf{p}}=1\text{m}^2$, $\sigma_{\mathbf{v}}=1\left(\text{m} / \text{s}\right)^2.$ Besides, in order to make simulations closer to the practice, the simulation step of the robot model is 0.001s, while the simulation steps of the controller and the noise generator are both 0.02s.

\begin{figure}
	\centering
	\includegraphics[width=\columnwidth]{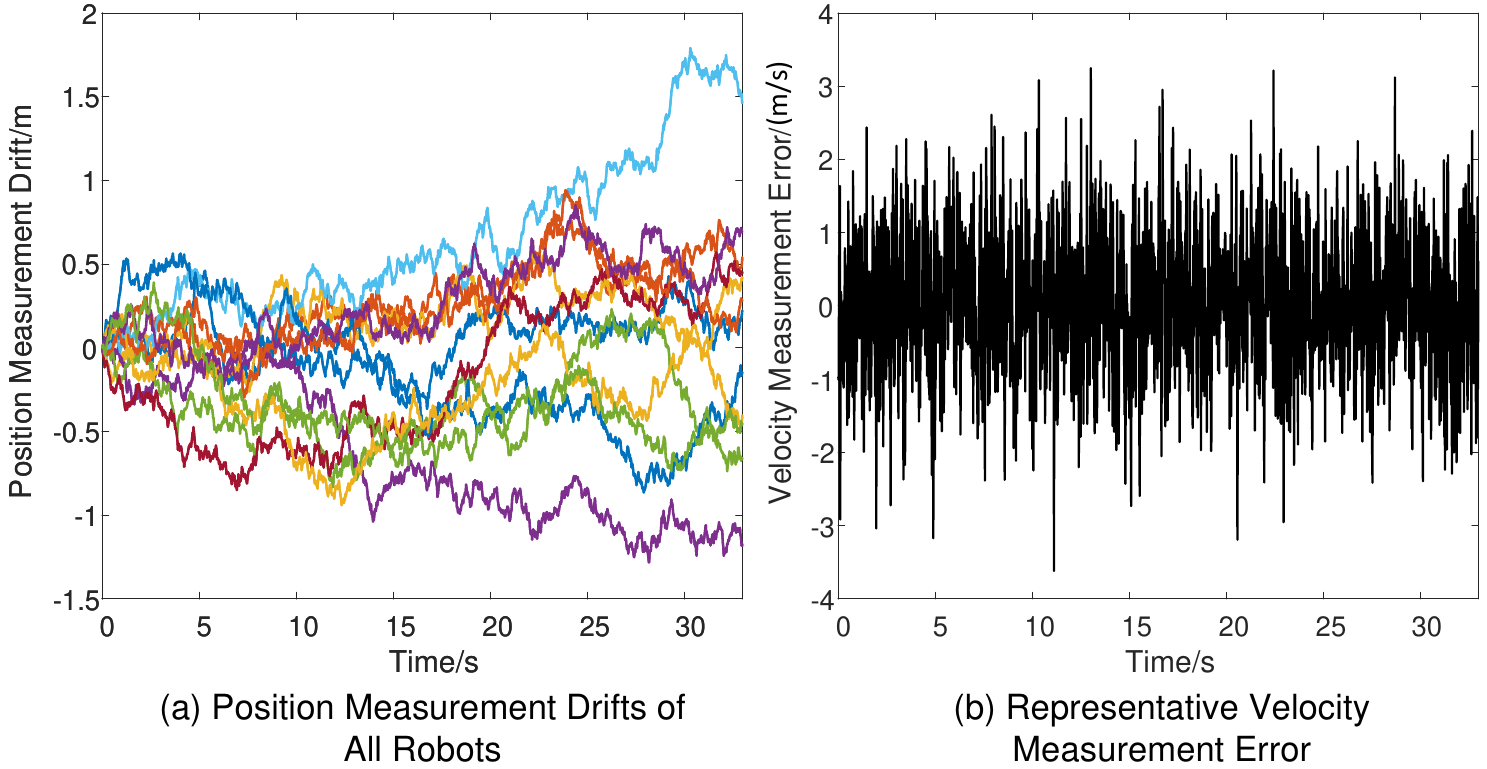}
	\caption{Position measurement drifts of all robots and a representative velocity measurement error.}
	\label{ReSimNoise}
\end{figure}

\begin{figure}
	\centering
	\includegraphics[width=\columnwidth]{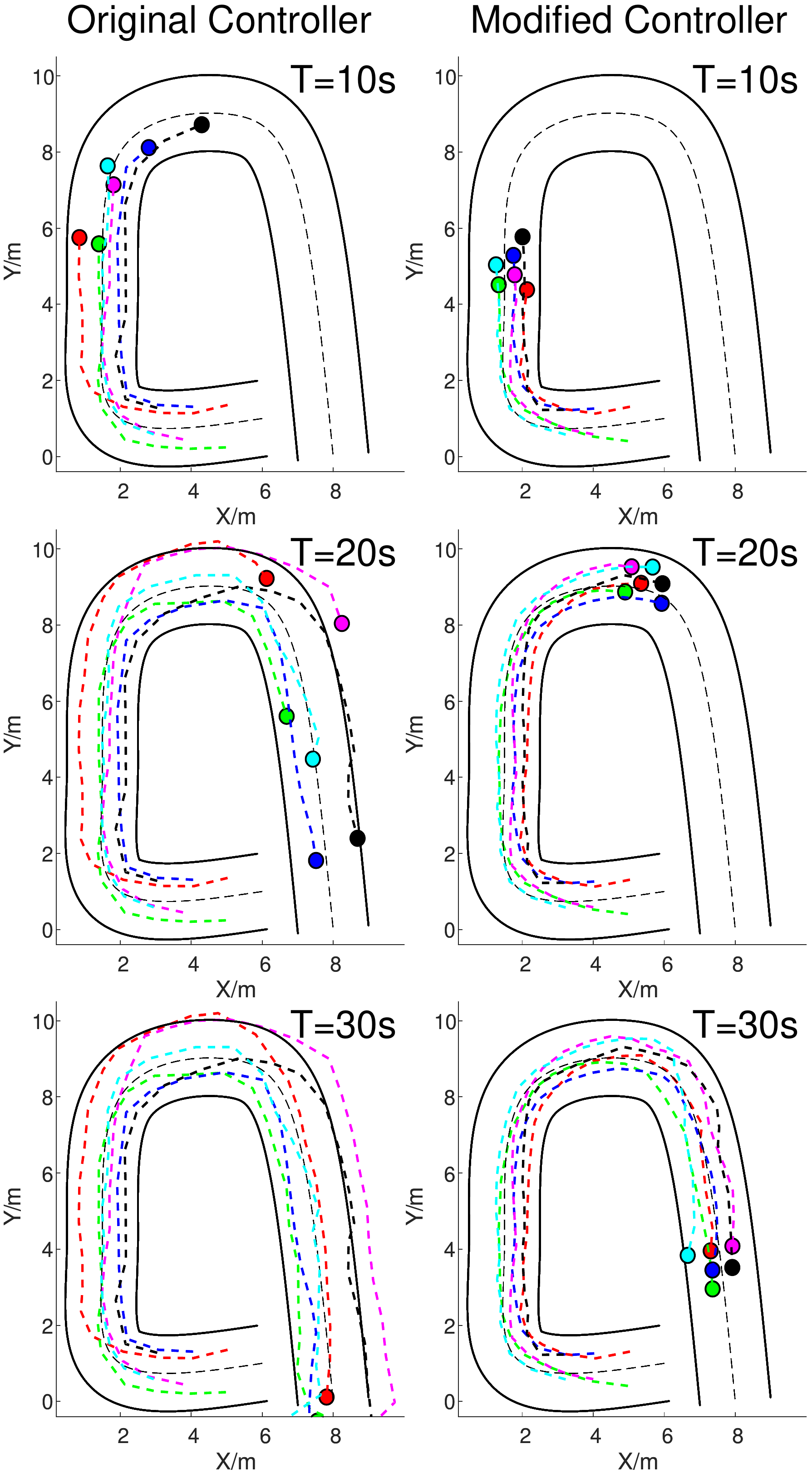}
	\caption{Snapshots of comparative simulations.}
	\label{ReSimCom}
\end{figure}
\begin{figure}
	\centering
	\includegraphics[width=\columnwidth]{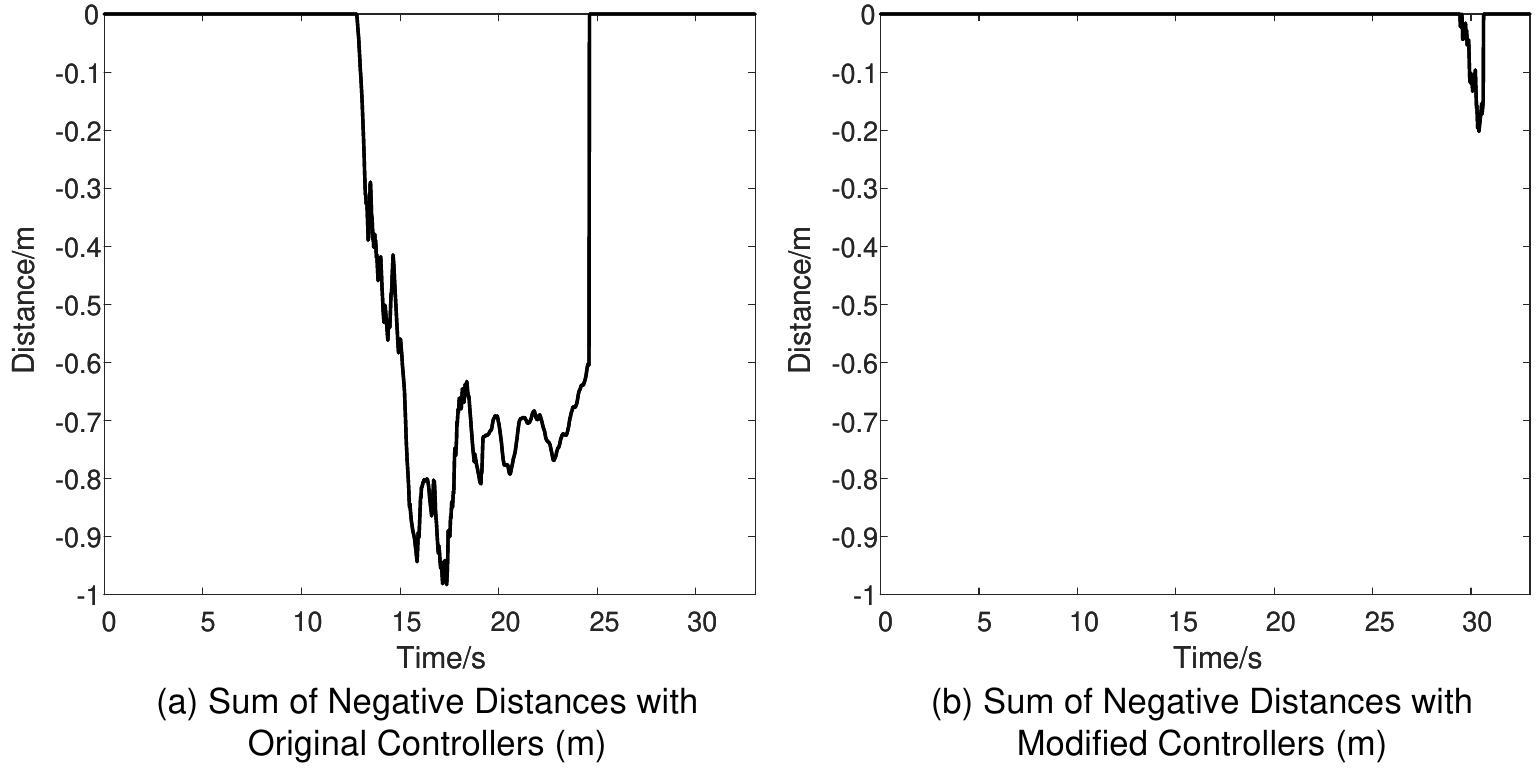}
	\caption{The comparison of the sum of negative distances $d_\text{t,all}$ from the tube boundary of all robots in the simulations.}
	\label{ReSimCurve}
\end{figure}

Both simulations last 33 seconds. During the simulation, the position measurement drifts of all robots and a representative velocity measurement error are shown in Figure \ref{ReSimNoise}. As shown in Figure \ref{ReSimCom}, three snapshots of each simulation are presented, and the safety areas of all robots are represented by circles with different colors. It can be observed that with original controllers \eqref{VelNoVA}, \eqref{AccNoVA}, robots often move outside $\mathcal{T}_{\mathcal{V}}$ during the simulation due to the existence of the self-localization drift. By comparison, robots usually keep moving inside $\mathcal{T}_{\mathcal{V}}$ with modified controllers \eqref{AccNoVA}, \eqref{vmdf}. 

For a better comparison of two kinds of controllers, a \emph{sum of negative distance} $d_\text{t,all}$ is proposed as the comparison index, which is defined as
\begin{equation*}
	d_\text{t,all}=\sum_{i=1}^{M}\min\left(d{_{\text{t,}i}}-r_{\text{s}},0\right),
\end{equation*}where $d{_{\text{t,}i}} = r_{\text{t}}\left( \mathbf{p}_{i}\right) - \left \Vert \mathbf{p}_{i}-\mathbf{m}\left( \mathbf{p}_i\right) \right \Vert$ represents the distance between the $i$th robot and the tube boundary. There always exists $d_\text{t,all}\leq 0$. A smaller $d_\text{t,all}$ means that the phenomenon of robots moving outside $\mathcal{T}_{\mathcal{V}}$ is more common. In Figure \ref{ReSimCurve}, a comparison about the index $d_\text{t,all}$ is proposed. It is obvious that modified controllers proposed in this paper perform better and have the ability to reduce the negative impact of the self-localization drift.

\subsection{Real Experiment}
In this subsection, real experiments are implemented on an open-source platform called Robotarium, which is developed by Georgia Institute of Technology \cite{pickem2017robotarium,wilson2020robotarium}. By uploading our algorithms into the Web front end, the control algorithms are verified by the user management and transferred to the experiment server. In the experiments, $M=10$ ground mobile robots are asked to move in a predefined curve virtual tube. Here the curve virtual tube is closed, namely the generating curve starts and ends at the same point. The width of $\mathcal{T}_{\mathcal{V}}$ is always $r_\text{t}=0.25\text{m}$ along the generating curve. The original controllers \eqref{VelNoVA}, \eqref{AccNoVA} and modified controllers  \eqref{AccNoVA}, \eqref{vmdf} are applied to guide these robots, respectively. The control parameters are $k_2=k_3=k_5=1$, $k_4=2$, $r_\text{s} = 0.075\text{m}$, $r_\text{a} = 0.125\text{m}$, $r_\text{c} = 0.3\text{m}$, $r_\text{d} = 1\text{m}$, $v_\text{m} = 0.1\text{m/s}$. The variances of noises $\mathbf{n}_{\mathbf{p}_i}$ and $\mathbf{n}_{\mathbf{v}_i}$ are set as $\sigma_{\mathbf{p}}=10^{-5}\text{m}^2$, $\sigma_{\mathbf{v}}=10^{-5}\left(\text{m} / \text{s}\right)^2$.

\begin{figure}
	\centering
	\includegraphics[width=\columnwidth]{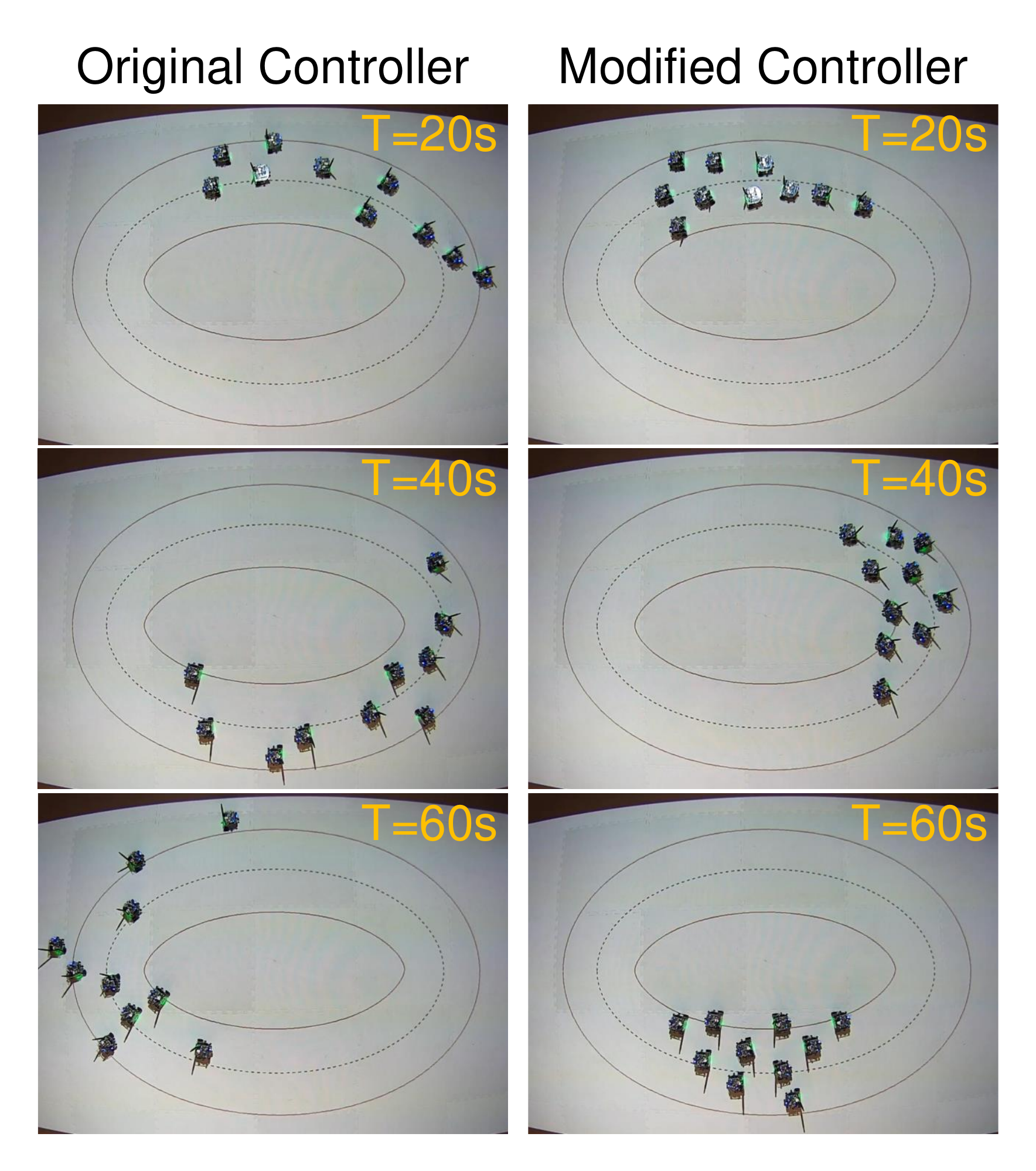}
	\caption{Snapshots of real experiments.}
	\label{ExperCom}
\end{figure}
\begin{figure}
	\centering
	\includegraphics[width=\columnwidth]{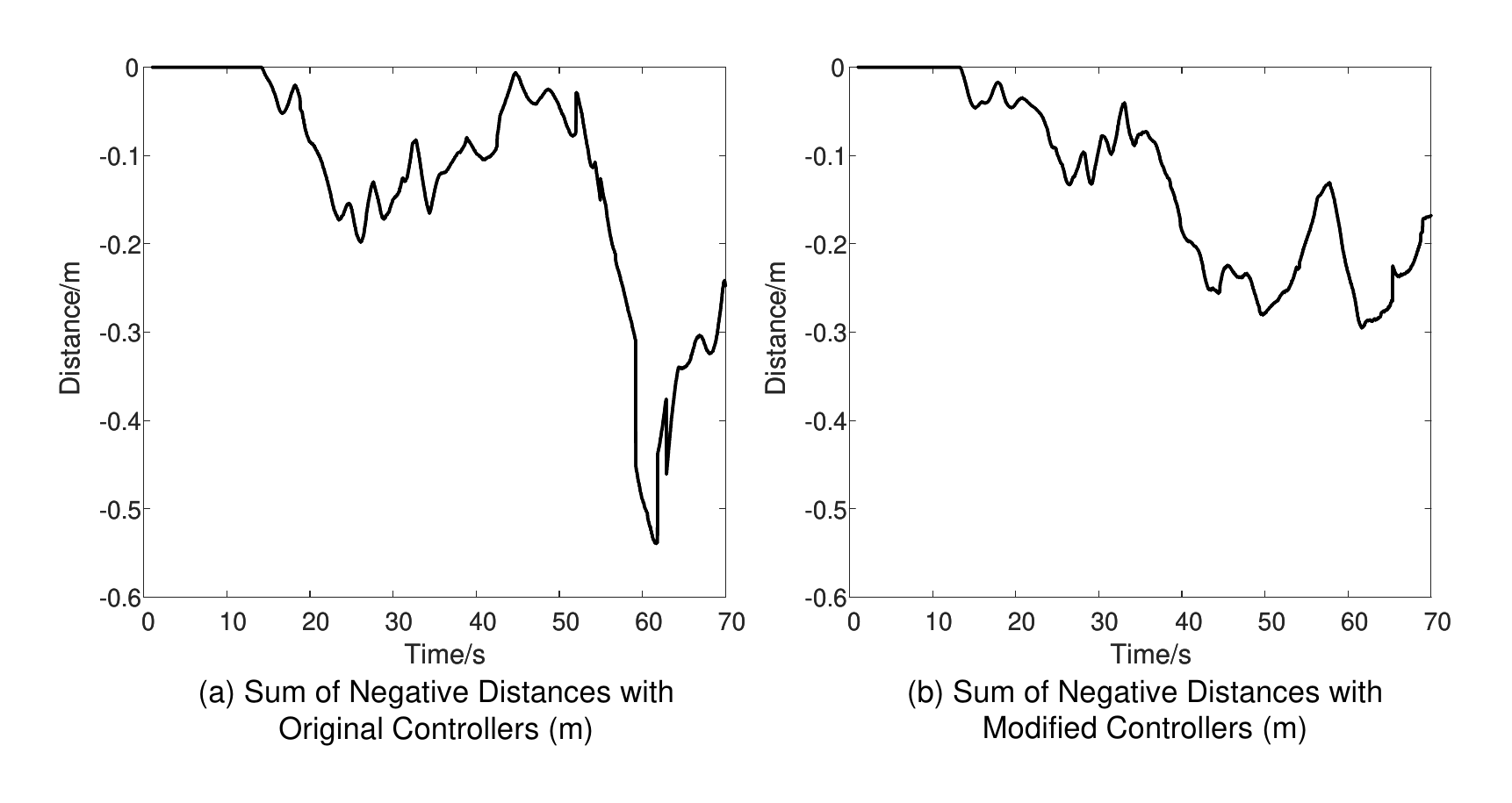}
	\caption{The comparison of the sum of negative distances $d_\text{t,all}$ from the tube boundary of all robots in the experiments.}
	\label{ExperCurve}
\end{figure}

Both experiments last 70 seconds. As shown in Figure \ref{ExperCom}, three snapshots of each experiment are presented. It can be observed that compared with original controllers \eqref{VelNoVA}, \eqref{AccNoVA}, robots have more tendencies to keep moving inside $\mathcal{T}_{\mathcal{V}}$ with modified controllers \eqref{AccNoVA}, \eqref{vmdf}. In Figure \ref{ReSimCurve}, the comparison about the index $d_\text{t,all}$ is proposed. It is obvious that modified controllers proposed in this paper perform better and have the ability to reduce the negative impact of the self-localization drift.

\section{Conclusions}

The robust curve virtual tube passing through problem is proposed and then solved in this paper. To reduce the negative impact of the self-localization drift, the flocking algorithm is introduced to the controller design. And a modified vector field controller is designed based on our previous work \cite{quan2021distributed}. Formal analyses and
proofs are made to show that the cohesion behavior and the velocity alignment behavior have the ability to reduce the influence of the position measurement drift and the velocity measurement error, respectively. 
Finally, comparative simulations and real experiments are given to show the effectiveness and performance of the proposed method.




\section*{Appendix}
\subsection{Proof of Lemma 2}
	 Firstly, for the acceleration command \eqref{AccNoVA}, the input and output are defined as $\mathbf{n}_{\mathbf{v}_i}$ and $\mathbf{v}_{i}$, respectively. Then a transfer function matrix $\mathbf{G}\left(s\right)$ is obtained as
	\begin{align*}
		\mathbf{G}\left(s\right)={G}\left(s\right)\mathbf{I}_2=\frac{k_\mathbf{v}}{s+k_\mathbf{v}}\mathbf{I}_2.	
	\end{align*}
	As $\mathbf{n}_{\mathbf{v}_i}$ is a two-dimensional white Gaussian noise, the  power spectral density of each dimension $s_{\mathbf{n}_{\mathbf{v}_i}}\left(\omega\right) $ is always equal to its variance, namely $s_{\mathbf{n}_{\mathbf{v}_i}}\left(\omega\right) =\sigma_{\mathbf{v}} $ \cite{grimmett2020probability}. Besides, as $\mathbf{G}\left(s\right)$ is linear and  time-invariant, there exists a two-dimensional zero mean white Gaussian noise component in $\mathbf{v}_{i}$ \eqref{AccNoVA} \cite{grimmett2020probability}. Then, it has
$
		\sigma_{\mathbf{v}}^{\prime}=\frac{1}{2\pi}\int_{-\infty}^{+\infty}\left\vert G\left(j\omega\right)\right\vert^2 s_{\mathbf{n}_{\mathbf{v}_i}}\left(\omega\right) \text{d}\omega  =\frac{\sigma_{\mathbf{v}}}{2\pi}\int_{-\infty}^{+\infty}\frac{k_\mathbf{v}^{2}}{\omega^2+k_\mathbf{v}^{2}} \text{d}\omega 
		=\frac{k_\mathbf{v}}{2}\sigma_{\mathbf{v}}.
$
	
	 Secondly, for the modified acceleration command \eqref{ModifiedAccelerationController}, the input is defined as $\mathbf{u}=\left[\mathbf{n}_{\mathbf{v}_1}^{\text{T}}\ \mathbf{n}_{\mathbf{v}_2}^{\text{T}}\cdots\ \mathbf{n}_{\mathbf{v}_N}^{\text{T}}\right]^{\text{T}}$, and the output is defined as $\mathbf{y}=\left[\mathbf{v}_{1}^{\text{T}}\ \mathbf{v}_{2}^{\text{T}}\cdots \ \mathbf{v}_{N}^{\text{T}}\right]^{\text{T}}$.
	The relative localization relationship among $N$ robots is modeled as a simple connected undirected graph. Then, the diagonal elements of the Laplacian matrix $\mathbf{L}\in\mathbb{R}^{N\times N}$ are all $N-1$, and other elements are all $-1$. Then the state equations are expressed as
$
		\dot{\mathbf{v}}=\mathbf{A}\mathbf{v}+\mathbf{B}\mathbf{u},
		{\mathbf{y}}=\mathbf{C}\mathbf{v},
$
	where $\mathbf{A}=-k_\mathbf{v}k_{5}\left(\mathbf{L}\otimes \mathbf{I}_2+\frac{1}{k_{5}}\mathbf{I}_{2N}\right)$, $\mathbf{B}=k_\mathbf{v}\mathbf{I}_{2N}$, $\mathbf{C}=\mathbf{I}_{2N}$.
	The transfer function matrix $\mathbf{H}\left(s\right)$ is shown as 
	\begin{align*}
		\mathbf{H}\left(s\right)
		=\left[
		\begin{matrix}
			H_1\left(s\right)\mathbf{I}_{2} & H_2\left(s\right)\mathbf{I}_{2}  & \cdots & H_2\left(s\right)\mathbf{I}_{2}   \\
			H_2\left(s\right)\mathbf{I}_{2}    & H_1\left(s\right) \mathbf{I}_{2}  & \cdots & H_2\left(s\right)\mathbf{I}_{2}  \\
			\vdots & \vdots & \ddots & \vdots \\
			H_2\left(s\right) \mathbf{I}_{2}  & H_2\left(s\right)\mathbf{I}_{2}   & \cdots & H_1\left(s\right)\mathbf{I}_{2} 
		\end{matrix}
		\right],
	\end{align*}
	where  
$
	H_1\left(s\right)=\frac{k_\mathbf{v}\left(s+k_\mathbf{v}+k_\mathbf{v}k_{5}\right)}{\left(s+k_\mathbf{v}\right)\left(s+k_\mathbf{v}+Nk_\mathbf{v}k_{5}\right)}, H_2\left(s\right)=\frac{k_\mathbf{v}^{2}k_{5}}{\left(s+k_\mathbf{v}\right)\left(s+k_\mathbf{v}+Nk_\mathbf{v}k_{5}\right)}.
$
	Hence, for the $i$th robot, there exists
$
		\mathbf{v}_{i}\left(s\right)=H_1\left(s\right)\mathbf{n}_{\mathbf{v}_i}+\sum_{j=1,j\neq i}^{N}H_2\left(s\right)\mathbf{n}_{\mathbf{v}_j}.
$	
	As $\mathbf{H}\left(s\right)$ is linear and time-invariant, there exists a two-dimensional zero mean white Gaussian noise component in $\mathbf{v}_{i}$ \eqref{ModifiedAccelerationController} \cite{grimmett2020probability}. Then, $\sigma_{\mathbf{v}}^{\prime\prime}$ is calculated as
$
		\sigma_{\mathbf{v}}^{\prime\prime}=\frac{1}{2\pi}\int_{-\infty}^{+\infty}\left\vert H_1\left(j\omega\right)\right\vert^2 s_{\mathbf{n}_{\mathbf{v}_i}}\left(\omega\right) \text{d}\omega +\sum_{j=1,j\neq i}^{N}\frac{1}{2\pi}\int_{-\infty}^{+\infty}\left\vert H_2\left(j\omega\right)\right\vert^2 s_{\mathbf{n}_{\mathbf{v}_i}}\left(\omega\right) \text{d}\omega.
$
	Then it has
	\begin{align*}
		&\sigma_\mathbf{v}^{\prime\prime}\!=\!\frac{\sigma_{\mathbf{v}}}{2\pi}\int\limits_{-\infty}^{+\infty}\left\vert H_1\left(j\omega\right)\right\vert^2  \text{d}\omega \!+\!\frac{\left(N-1\right)\sigma_{\mathbf{v}}}{2\pi}\int\limits_{-\infty}^{+\infty}\left\vert H_2\left(j\omega\right)\right\vert^2  \text{d}\omega \\
		&=\frac{\sigma_{\mathbf{v}}k_\mathbf{v}}{2}\frac{\left(N-1\right)\left(k_{5}N+k_{5}+2\right)+\left(k_{5}+2\right)\left(k_{5}N+1\right)}{N\left(k_{5}N+1\right)\left(k_{5}N+2\right)}\\ &\!+\!\frac{\left(N-1\right)\sigma_{\mathbf{v}}k_\mathbf{v}}{2}\frac{\left(N-1\right)k_{5}^2}{\left(k_{5}N+1\right)\left(k_{5}N+2\right)}
		\!=\!\frac{k_\mathbf{v}}{2}\frac{k_{5}+1}{k_{5}N+1}\sigma_{\mathbf{v}}. \square
	\end{align*}
	

\subsection{Proof of Theorem 2}
Under the ideal condition, $\mathbf{u}_{4,i}$ in \eqref{vmdf} has no influence on the system stability and safety as stated in \textit{Remark 1}. Similarly, $\mathbf{u}_{5,i}$ in \eqref{vmdf} also has no influence. The reason is that $\mathbf{u}_{5,i}$ can be equivalent to the derivative of a \emph{limited} potential function, which is easy to add to the final Lyapunov-like function as stated in \cite{quan2021distributed}. 
Then, according to \textit{Theorem 1} and \textit{Lemma 1}, there exists $t_{1}>0 $ such that all robots can pass through  $\mathcal{T}_{\mathcal{V}}$ at $t\geq t_{1},$ meanwhile guaranteeing $\mathcal{S}_{i}\left(t\right)\cap \mathcal{S}_{j}\left(t\right)=\varnothing,$ $\mathcal{S}_{i}\left(t\right)\cap \partial \mathcal{T}_{\mathcal{V}}=\varnothing$, $t\in \lbrack 0,\infty ),i,j=1,2,\cdots,M,i \neq j$.

Next consider the condition that \textit{Assumption 1} is satisfied. When the saturation function in \eqref{vmdf} does not work, the controllers \eqref{AccNoVA}, \eqref{vmdf} can make the robotic swarm pass through $\mathcal{T}_{\mathcal{V}}$ more safely and more precisely according to \textit{Lemma 2}. When the saturation function in \eqref{vmdf} works, the norms of five control terms in \eqref{vmdf} scale proportionally. The control effect of the robot cohesion and velocity alignment may become weaker. But modified controllers can still make the robotic swarm pass through $\mathcal{T}_{\mathcal{V}}$ more safely and more precisely. $\square$


\bibliographystyle{IEEEtran}
\bibliography{RobustCurveTube}

%

\end{document}